\documentclass[journal]{IEEEtran}
\pdfoutput=1
\ifCLASSINFOpdf
\else
\fi

\usepackage[english]{babel}
\usepackage{blindtext}
\usepackage[titlenumbered,ruled]{algorithm2e}
\usepackage{mathtools}          
\usepackage[pdftex]{graphicx}
\usepackage{bbm}
\usepackage{multirow}
\usepackage{pdflscape}
\usepackage{afterpage}
\usepackage{stfloats}
\begin{document}
	%
	
	
	
	\title{Dynamics of Driver's Gaze: Explorations in Behavior Modeling \& Maneuver
		Prediction}
	
	%
	%
	%
	
	\author{Sujitha~Martin,~\IEEEmembership{Member,~IEEE,}
		Sourabh~Vora,~
		Kevan~Yuen,~
		and~Mohan~M.~Trivedi,~\IEEEmembership{Fellow,~IEEE}
		\thanks{The authors are with the Laboratory for Intelligent and Safe
			Automobiles,
			University of California San Diego, La Jolla, CA 92093 USA: (see
			http://cvrr.ucsd.edu/).}
	}
	
	%
	%

	\markboth{}
	{Shell \MakeLowercase{\textit{Martin et al.}}: Vision based Gaze-dynamics
		Modeling, and Behavior Understanding and Prediction in Naturalistic Driving}
	%



	\maketitle
	
	\begin{abstract}
		The study and modeling of driver's gaze dynamics is important because, if and how the driver 
		is monitoring the driving environment is vital for driver assistance in manual mode, 
		for take-over requests in highly automated mode and for semantic perception of the 
		surround in fully autonomous mode. We developed a machine vision based framework 
		to classify driver's gaze into context rich zones of interest and model driver’s gaze behavior
		by representing gaze dynamics over a time period using gaze accumulation, glance
		duration and glance frequencies. As a use case, we explore the driver's gaze dynamic
		patterns during maneuvers executed in freeway driving, namely, left lane change
		maneuver, right lane change maneuver and lane keeping. It is shown that condensing
		gaze dynamics into  durations and frequencies leads to recurring patterns
		based on driver activities. Furthermore, modeling these patterns show predictive
		powers in maneuver detection up to a few hundred milliseconds a priori.
	\end{abstract}
	

	%
	\IEEEpeerreviewmaketitle

	\section{Introduction}
	%
	%
	%
	%
	\IEEEPARstart{I}{ntelligent} vehicles of the future are that which, having a 
	holistic perception (i.e. inside, outside and of the vehicle) and understanding 
	of the driving environment, make it possible for occupants to go from point A 
	to point B safely, comfortably and in a timely manner \cite{TrivediITS2007,DoshiITSC2011}. This may happen with the 
	driver in full control and getting active assistance from the robot, or the robot 
	is in partial or full control and human drivers are passive observers ``ready'' to 
	take over as deemed necessary by the machine or human \cite{SAE_J3106,CasnerCACM2016}. 
	In the full spectrum from manual to autonomous mode, modeling the dynamics of 
	driver's gaze is of particular interest because, if and how the driver is monitoring the
	driving environment is vital for driver assistance in manual mode \cite{JainICCV2015}, for take-over
	requests in highly automated mode \cite{GoldHFESAM2013} and for semantic perception of the surround in
	fully autonomous mode \cite{TawariIV2017,PalazziIV2017}. 
	
	\begin{figure}
		\centering
		\includegraphics[width=0.90\linewidth]{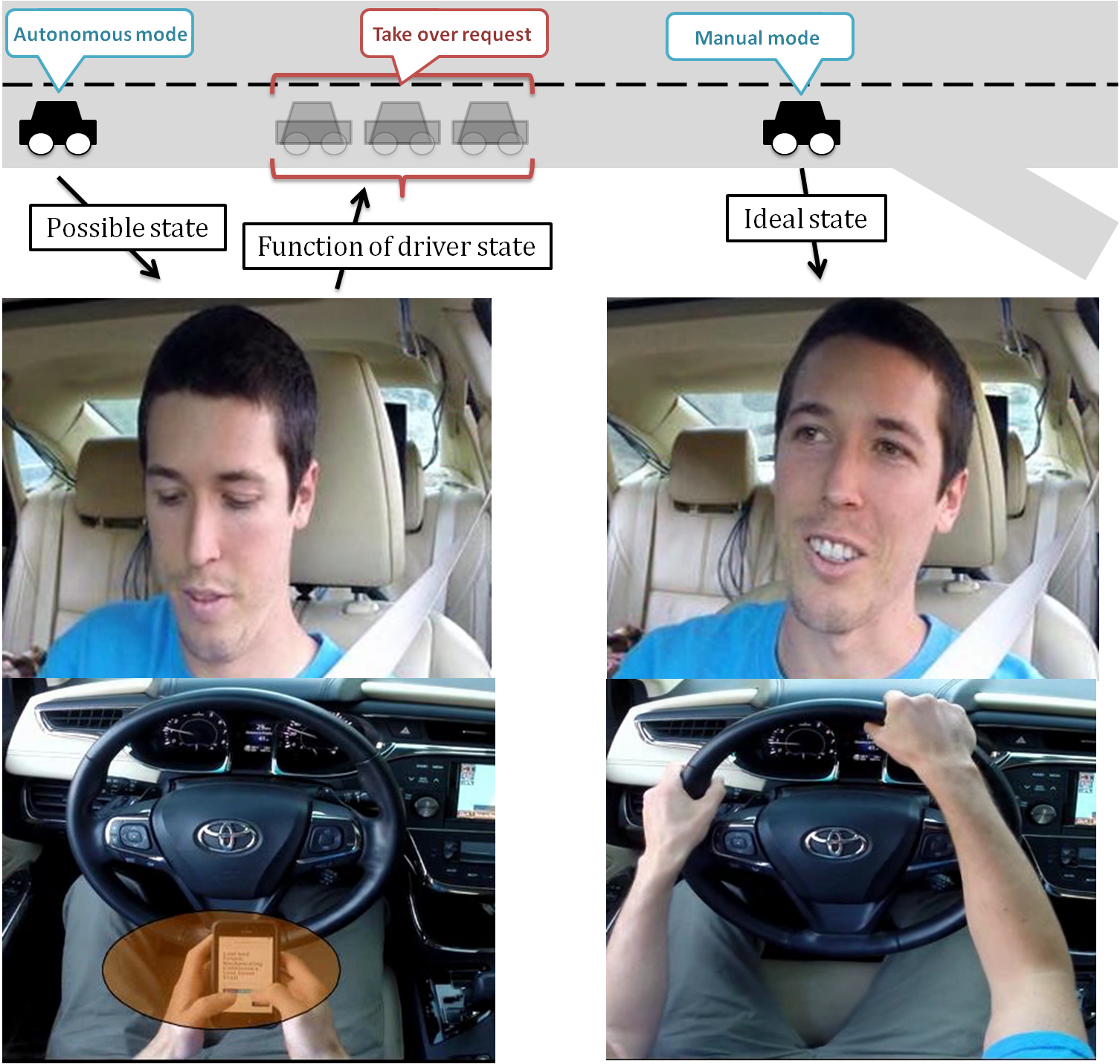}
		\caption{
			An example illustration which showcases the importance of understanding and modeling what constitutes expected or "attentive" gaze behavior in order to ensure safe and smooth transfer of control between robot and human.}
		\label{fig:motivation}
	\end{figure}
	
	The driver's gaze can be represented in many different ways, from directional vectors \cite{VicenteITSS2015} to points in 3-D space \cite{ZhangCVPRW2017}, from static zones of interest (e.g. speedometer, side mirrors) \cite{TawariChenITSC2014} to dynamic objects of interest (e.g. vehicles, pedestrians) \cite{TawariITSC2014}. In this paper, gaze is represented using context rich static zones of interest. Using such a representation, when gaze is estimated over a period of time, higher semantic information such as fixations and saccades can be extracted and  used to derive driver's situational awareness, estimate engagement in secondary activities, predict intended maneuvers, etc. Figure \ref{fig:motivation} illustrates an example where the length of driver's fixation on non-driving relevant region is an important factor to determine driver's state and therefore, when the transfer of control should happen. 
	
	In general, with rapid introductions of autonomous features in consumer vehicles, there is a need to understand and model what constitutes ``normal'' or ``attentive'' gaze behavior in order to ensure safe and smooth transfer of control between robot and human. An important criteria to build such models is naturalistic driving data, where driver is in full control. Using such data, we propose to build expected gaze behavior models for a given situation or activity and attempt to predict the presence or absence of such behavior on data unseen when training the models. For example, if we build a gaze model from left lane change events alone, then applying the model to new unlabeled data gives the likelihood of a left lane change event occurring or more abstractly, driver's situational awareness necessary to make a left lane change. One of the challenges is in the mapping from spatio-temporal rich gaze dynamics to activities or events of interest. For example, when engaged in a secondary task which uses the center stack (e.g. radio, AC, navigation), the manner in which the driver looks at the center stack can be vastly different. One may perform the secondary task with one long glance away from the forward driving direction and at the center stack. Another time one may perform the secondary task via multiple short glances towards the center stack, etc. However, while individual gaze dynamic patterns are different, together they are associated with an activity of interest.

	To this end, we present a machine vision based framework focused on gaze-dynamics
	modeling and behavior prediction using naturalistic driving data. Whereas a preliminary 
	study of this work using a single driver is presented in \cite{MartinIV2017}, the 
	contributions of this paper are as follows:
	\begin{itemize}
		\item A significant overview of related studies on gaze estimation and higher semantics with gaze.
		\item Naturalistic driving dataset composed of multiple drivers, and annotated with ground truth gaze zones and maneuver execution (e.g. lane change, lane keeping).
		\item New metrics to quantitatively evaluate the performance of gaze estimation over a time period as oppose to on individual frame level (i.e. metrics to evaluate gaze accumulation which is computed over a time segment).
		\item Proper formulation and nomenclature of gaze-dynamic features (i.e. gaze accumulation, glance duration, glance frequency), and compare-and-contrast on the effect of utilizing a combination of these features on behavior prediction accuracy.
		
	\end{itemize}

	\section{Related Studies}
	The work presented in this paper has three major components: gaze estimation,
	gaze behavior modeling and prediction, and performance evaluation. Table \ref{table:relatedWorks} 
	reflects these attributes by dividing the columns
	into two major sections, gaze estimation and higher semantics with gaze. Some
	works present automatic gaze estimation frameworks but don't proceed further,
	while some use manually annotated gaze data to study higher semantics. In the
	two studies which present work in both categories, the differences are subtle
	but significant; first is the number of gaze zones, second is in the features
	used for behavior modeling and prediction, third is in the performance
	evaluation of gaze zones and behavior prediction. Note that the studies
	presented in Table I are selected based on two
	criteria: first, it must present work on driver's gaze and second, evaluation is
	conducted on some level with on-road driving data.

	
	\subsection{On Gaze Estimation}
	
	In literature, works on gaze zone estimation are relatively new and of those,
	there are two categories: geometric and learning based methods. 
	The work presented in \cite{AhlstromITSS2013} estimates gaze zones
	based on geometric methods, where a 3-D model car is divided into different zones
	and 3D gaze tracking is used to classify gaze into predefined zones; however, no
	evaluations on gaze zone level is given. Another
	geometric based method is presented in \cite{VicenteITSS2015}, but the number of
	gaze zones estimated is very limited (i.e. on-road versus off-road) and
	evaluations are conducted in stationary vehicles. In terms of learning based
	methods, there are two prevalent works. Work by Tawari et al.
	\cite{TawariChenITSC2014} has the most similarity to the work presented in this
	paper in terms of the features selected (e.g. head pose, horizontal gaze surrogate), 
	classifier used (i.e. random forest)
	and evaluation on naturalistic driving data. The difference is that this work
	introduces another feature to augment the state of the eyes (i.e. appearance
	descriptor), which allows for an increased number of gaze zones, but not at the 
	expense of performance, as shown by evaluating on a dataset composed of multiple drivers. 
	Another learning based method is the work presented by Fridman
	et al. \cite{FridmanIETCV2016} where the evaluations are done on
	a significantly large dataset, but the design of the features to represent
	the state of the head and eyes are what is causing their classifier to over fit
	to user based models and to not generalize well with global based models.
	
	 
	
	\afterpage{%
		\begin{landscape}
			\centering
			{\footnotesize
				\begin{table}
					\caption{Selected studies on vision based gaze estimation and higher
						semantics with gaze which are evaluated on some level with on-road driving
						data.}
					\begin{tabular}{|p{2cm}|p{3.3cm}|p{1.5cm}|p{1.1cm}|p{1.4cm}|p{2.5cm}|p{1.7cm}|p{1.7cm}|p{2.5cm}|p{2cm}|}
\hline
& & \multicolumn{4}{|c|}{Gaze Estimation} & \multicolumn{4}{|c|}{Higher Semantics with Gaze} \\
\cline{3-10}
& & & Num. of & \multicolumn{2}{|c|}{Evaluation}&&&&\\
\cline{5-6}
Research Study & Objective / Motivation & Methodology & Gaze & Over conti- & Accuracy&Features & Method & Behavior / task / state & Prediction\\
& && Zones & -nuous time  &&&& of interest & accuracy \\
\hline
\hline
Tawari et al., 2014 \cite{TawariITSC2014} & Estimating driver attention by simultaneous analysis of viewer and view & Geometric & Function of salient objects & No & 46\% and 79\% with manual and automatic detection, respectively, of salient objects & Not applicable & Not applicable & Not applicable & Not applicable \\
\hline
Tawari, Chen \& Trivedi, 2014 \cite{TawariChenITSC2014} & Estimate driver’s coarse gaze direction using both head and eye cues &Learning & 6 & No & 80\% with head pose alone and 95\% with head plus eye cues & Not applicable & Not applicable & Not applicable & Not applicable\\
\hline
Vicente et al., 2015 \cite{VicenteITSS2015} & Detecting eyes off the road (EOR) & Geometric & 2 & No & 90\% EOR accuracy & Not applicable & Not applicable & Not applicable & Not applicable \\
\hline
Vasli, Martin \& Trivedi, 2016 \cite{VasliITSC2016} & Exploring the fusion of geometric and data driven approaches on driver gaze estimation & Geometric plus learning & 3 & No &  75\% with geometric and 94\% with geometric plus learning & Not applicable & Not applicable & Not applicable & Not applicable \\
\hline
Fridman et al., 2016 \cite{FridmanIETCV2016} & Exploring the effects of head pose and eye pose on gaze & Learning & 6 & No & 89\% with head pose alone and 95\% with head and eye pose & Not applicable & Not applicable & Not applicable & Not applicable \\
\hline
\hline
Birrell \& Fowkes, 2014 \cite{BirrellTRP2014} & Investigates glance behaviors of drivers when using Smartphone application & Manual annotation & 8 & No & Not applicable & GL, GD, GF & Not applicable & Secondary task versus baseline driving & Not applicable \\
\hline
Munoz et al., 2016 \cite{MunozTRPF2016} & Predicting tasks based on distinguishing patterns in driver's visual attention allocation & Manual annotation & 11 & No & Not applicable & GL, GD & HMM & Secondary tasks & Min of 68\% to max of 96\% \\
\hline
Fridman et al., 2016 \cite{FridmanHFCS2016} & Exploring what broad macro eye-movement reveal about state of driver and  driving environment & Manual annotation & 8 & No & Not applicable & GD, GTF & HMM & Driving environment, driver behavior/state, driver demographic characteristic & Min of 52\% to max of 88\% \\
\hline
\hline
Ahlstrom, Kircher \& Kircher, 2013 \cite{AhlstromITSS2013}& Investigate the usefulness of a real-time distraction detection algorithm called AttenD & Geometric & Not available & No & Not available & GL, GD & Rule based & Attention to field relevant to driving & Not available \\
\hline
Li \& Busso, 2016 \cite{LiTITS2016} & Detecting mirror checking actions and its application to maneuver and secondary task recognition & Learning & 2 & No & Using all features from CAN, road cam and face cam: 90\% weighted and 96\% unweighted accuracy & GL, GD, GF, CAN-Bus signal, road dynamics &LDC & Vehicle maneuvers, secondary tasks & Min of 58\% to max of 76\% \\
\hline
This work & Estimating gaze dynamics and investigating the predictive power of glance duration and frequency on driver behavior & Learning & 9 & Yes & 84\% weighted accuracy and mostly above 25\% in ratio of estimated to true gaze accumulation & GL, GA, GD, GF & MVN & Left/right lane changes, lane keeping & Min of 78\% and max of 84\% \\
\hline
\multicolumn{10}{c}{GA = gaze accumulation, GL = glance location,	GD = glance duration,	GF=Glance frequency, HMM = Hidden Markov Model, LDC = Linear Discriminant Classifier, MVN = Multivariate Normal}
\end{tabular}
					\label{table:relatedWorks}
				\end{table}
			}
			\clearpage
		\end{landscape}
	}
	
	\subsection{On Gaze Behavior}


%
%
%
	
	In terms of gaze modeling and behavior understanding, literary works have mainly
	conducted studies in a driving simulator but few recent works from on-road driving have
	emerged. In one on-road study, Birrell and Fowkes \cite{BirrellTRP2014} explore
	the effects of using in-vehicle smart driving aid on glance behavior. The study 
	uses glance durations and glance transition	frequencies to show difference in 
	glance behavior between baseline, normal driving and when using in-vehicle devices. 
	Similarly, through manual annotations of glance times and targets, Munoz et al. 
	\cite{MunozTRPF2016} analyzed glance allocation strategies experimentally under 
	three different situations, manual radio tuning, voice-based radio tuning and 
	normal driving.
	In another on-road study, Li et al. \cite{LiTITS2016} show that
	drivers exhibit different gaze behaviors when engaged in secondary tasks versus
	baseline, normal driving by using mirror-checking actions as indicators for
	differentiating between the two. In addition to gaze related features, 
	Li et al. also employed features from CAN-Bus and road camera when training to detect
	mirror-checking actions, which raises a question of if the system is actually learning what
	the driver should be doing rather than what the driver is doing. While most of the gaze
	behavior studies have largely centered on detecting the driver's state from driver's 
	glance allocation strategy, \cite{FridmanHFCS2016} goes beyond to ask whether external
	driving environment can be inferred from six seconds of driver glances.


	
	
	
	
	In Table \ref{table:relatedWorks}, the gaze behavior related literature is divided into two different categories based on whether gaze estimation was performed automatically or manually. Such a distinction is presented in order to acknowledge works that have taken into consideration the noise in gaze estimates when modeling or predicting driver behavior from gaze. Our work especially addresses the effects of noisy gaze estimates on gaze behavior modeling by quantitatively evaluating gaze dynamic features (see Section V.B), as indicated by column four in Table \ref{table:relatedWorks}.

	

	
	\section{From Gaze Estimation to Dynamics \\ to Behavior Modeling}
	In this section, methods related to vision based gaze estimation, spatio-temporal rich gaze dynamics descriptors and behavior prediction from gaze modeling are described. 
	
	\subsection{Gaze Estimation}
	\label{sec:gaze_estimation}
\begin{figure}[!t]
		\centering
		\includegraphics[width=0.90\linewidth]{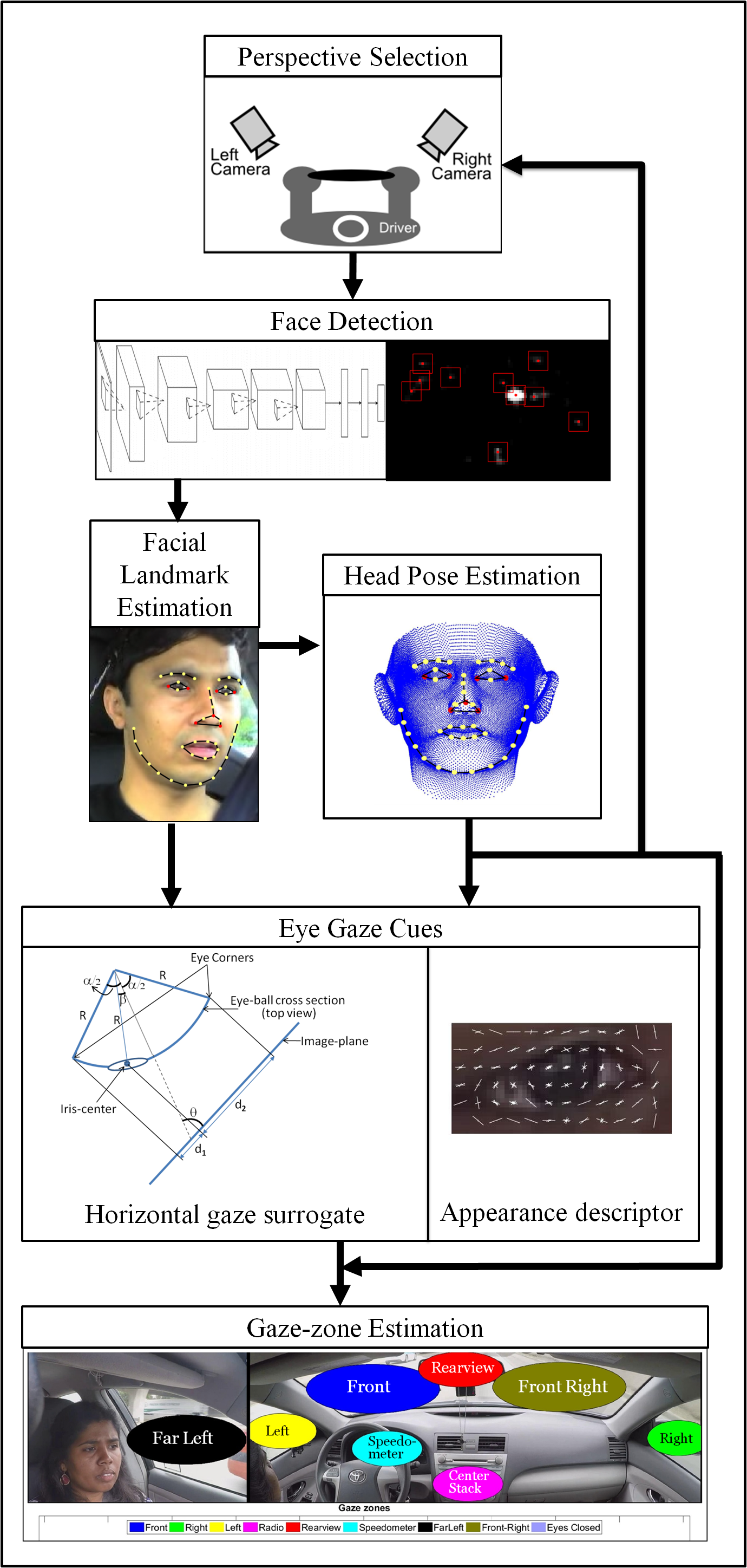}
		\caption{A illustrative block diagram showing the process of estimating gaze
			zone from time of capture from multiple camera perspectives to classifying gaze
			into one of nine gaze zones (i.e. eight gaze zones illustrated above and ``eyes
			closed''). Key modules in the system include deep CNN based face detection \cite{YuenICPR2016}, landmark estimation \cite{XiongCVPR2013}, horizontal gaze surrogate
			\cite{TawariChenITSC2014}, appearance descriptor and head pose based perspective selector \cite{MartinITSC2013}.}
		\label{fig:c02-gazezones}
	\end{figure}

	Gaze estimation is an important first step towards building gaze behavior models. As the emphasis of this work is on gaze behavior understanding, modeling and prediction, this work does not seek to claim major contribution in the domain of gaze estimation. However, for the sake of self-containment, this section provides high level information on the modules making up the gaze estimator and relevant references for more details. Key modules in this vision based gaze estimation framework, as illustrated in Figure \ref{fig:c02-gazezones}, are as follows:
	
%
%
	
	
	

	\begin{itemize}
		\item \textbf{Perspective Selection:} A part hardware and part software
		solution of distributed multi-perspective camera system, where each perspective
		is treated independently and a perspective is selected based on the dynamics and
		quality of head pose; details on head pose estimation is given below. Such a
		system is necessary to continuously and reliably track the head pose of driver during
		large head movements \cite{MartinITSC2013}.
		\item \textbf{Face Detection:} A deep CNN based system (with AlexNet as the base network) is trained on heavily augmented face datasets to include more
		examples of faces under harsh lighting and occlusion \cite{YuenICPR2016}.  
	\end{itemize}
	\begin{itemize}	
		\item \textbf{Facial landmark estimation:}  The landmarks are estimated using a
		cascade of regression models as described in \cite{BurgosICCV2013,XiongCVPR2013}
		with more details for iris localization given in \cite{TawariChenITSC2014}. 
		\item \textbf{Head pose estimation:} A geometric method where local features, such as eye corners,
		nose corners, and the nose tip, and their relative 3-D configurations,
		determine the pose \cite{MartinITSC2013}.
		
		\item \textbf{Horizontal gaze surrogate:} The horizontal gaze-direction $\beta$
		with respect to head, see Figure \ref{fig:c02-gazezones}, is estimated as a
		function of $\alpha$, angle subtended by an eye in horizontal direction,
		head-pose (yaw) angle $\theta$ with respect to the image plane, and
		$\frac{d_1}{d_2}$, the ratio of the distances of iris center from the detected
		corner of the eyes in the image plane \cite{TawariChenITSC2014}.
		\item \textbf{Appearance descriptor:} Appearance of the eye is represented by
		computing HoG (Histogram of Gradients) \cite{DalalCVPR2005} in a 2-by-2 patch around the eye. This
		descriptor is especially designed to capture the vertical gaze of the eyes.
		\item \textbf{Gaze zone estimation:} Eight semantic gaze-zones of interest are,
		\textit{far left}, \textit{left}, \textit{front}, \textit{speedometer},
		\textit{rear view}, \textit{center stack}, \textit{front right} and
		\textit{right}, as illustrated in Figure \ref{fig:c02-gazezones}. Another class
		of interest, but not illustrated in the figure, is the state of eyes closed.
		Consider a set of feature vectors $\vec{F} = \{\vec{f}_1, \vec{f}_2, \dots,
		\vec{f}_N\}$, and their corresponding class labels $X = \{x_1, x_2, \dots,
		x_N\}$, for $N$ sample instances. Here a feature vector is a concatenation of head pose
		and eye cues described above and class labels are one of nine gaze zones. Given
		$\vec{F}$ and $X$, a random forest (RF) is trained on the corpus. 
	\end{itemize}

	\begin{figure*}[!t]
		\centering
		\begin{tabular}{cc}
			
			\includegraphics[width=0.45\linewidth]{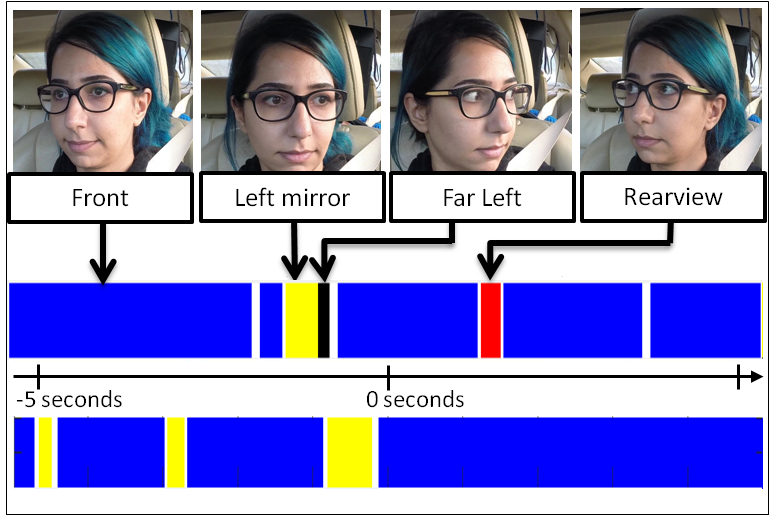}
			& 
			\includegraphics[width=0.45\linewidth]{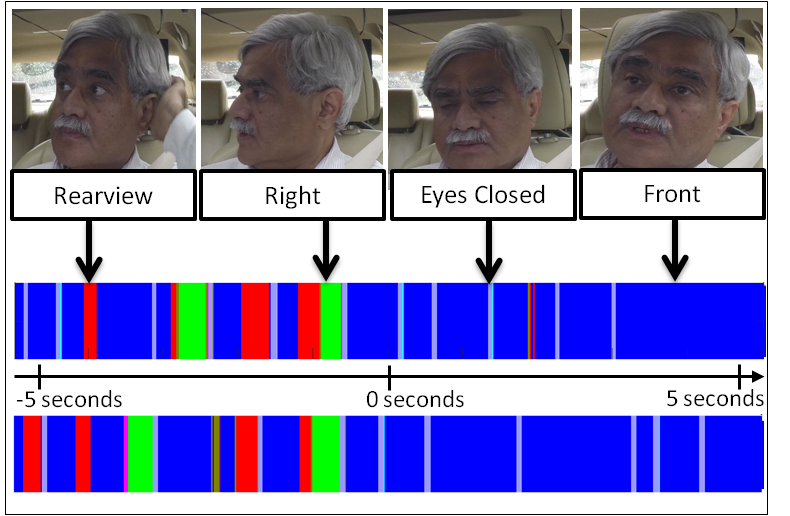}\\
			(a) Two Left Lane Change Events & (b) Two Right Lane Change Events\\
		\end{tabular}
		\caption{Illustrates four different scanpaths during a 10-second time window
			prior to lane change, two scanpaths during left lane change and two scanpaths
			during right lane change event, with sample face images from various gaze zones.
			Consistencies such as total glance duration and number of glances to regions of
			interest within a time window are useful to note when describing the nature of
			driver's gaze behavior. Such consistencies can be used as features to predict
			behaviors. See Figure \ref{fig:c02-gazezones} for a legend of which color is associate with which gaze zone.}
		\label{fig:c03-scanpaths}
		\vspace{-5mm}
	\end{figure*}

	\subsection{Spatio-Temporal Feature Descriptor}
	\label{sec:method_spatio_temporal_feature}
	The gaze estimator, as described in the previous section,
	outputs where the driver is looking in a given instance. A
	continuous segment of gaze estimates of where the driver
	has been looking is referred to as a scanpath. 
	Figure \ref{fig:c03-scanpaths} illustrates multiple scanpaths in a 10-second
	time window around lane change, two scanpaths from left lane
	change and two scanpaths from right lane change events. In the figure, the
	\textit{x-axis} represents time and the color displayed at a given time $t$
	represents the estimated gaze zone. Let \textit{SyncF} denote the time when the
	tire touches the lane marking before crossing into the next lane, which is the
	"0-seconds" displayed in the figure. Visually, in the 5-second time period
	before \textit{SyncF}, there is some consistency observed across the different
	scanpaths within a given event (e.g. left lane change); consistencies such as
	the minimum glance duration in relevant gaze zones. For example, in the scanpaths
	associated with right lane change, the driver glances at the \textit{rearview}
	and \textit{right} gaze zones for a significant duration. However, the start and
	end point of the glances are not necessarily the same across the different
	scanpaths. Therefore, we represent the scanpaths using features called gaze
	accumulation, glance frequency and glance duration, which remove some temporal
	dependencies but still capture sufficient spatio-temporal information to
	distinguish between different gaze behaviors.
	
	As these features are computed over a time window, we, first, define signals
	necessary to compute them. Let $Z$ represent the set of all nine gaze zones as
	$Z=$ \{\textit{Front}, \textit{Right}, \textit{Left}, \textit{Center Stack},
	\textit{Rearview}, \textit{Speedometer}, \textit{Left Shoulder}, \textit{Right
		Windshield}, \textit{Eyes Closed}\} and let $L=|Z|$ represent the total number
	of gaze zones. Let the vector $G=[g_1,g_2,\dots,g_N]$ represent the
	estimated gaze for an arbitrary time period of $T$, where $N =
	\textit{fps(frames per second)} \times T$, $g_n \in Z$, and $n \in
	\{1,2,\dots,N\}$. The following description defines how to compute gaze accumulation, glance duration and glance frequencies given $G$.
	
	\subsubsection{Gaze Accumulation}
	Gaze accumulation is a vector of size $L$, where each entry is a function of a unique gaze zone. Given a gaze zone, gaze accumulation is the accumulated sum of the number of times driver looked at the zone of interest within a time period; which is then normalized by the time window for relative accumulation.
	Mathematically, gaze accumulation at gaze zone $z_j$, where $j \in \{1,2,...,L\}$ corresponds to the $j^{th}$ gaze zone in $Z$, is as follows:
	\[
	\textit{Gaze Accumulation }(z_j) = \frac{1}{N} \times \sum_{n=1}^{N}
	\mathbbm{1}(g_n==z_j)
	\]
	where $\mathbbm{1}(\bullet)$ is the indicator function.
	%
	%
	%
	%
	%
	
	\subsubsection{Glance frequency}
	Glance frequency is a vector of size $L$, where each entry is a function of a unique gaze zone. Within time period $T$, every time there is a transition from one gaze zone to another (e.g. Front to Speedometer), the glance count for the destination gaze zone is incremented; the count is then normalized by the time period to produce glance frequency. 
	Under the condition that estimates are noise free, glance frequency for each of the gaze zones, $z_j$, where $j \in \{1,2,...,L\}$ corresponds to the $j^{th}$ gaze zone in $Z$, is calculated as follows:
	\begin{equation*}
		\begin{aligned}
			\textit{Glance } & \textit{Frequency } (z_j) \\
			& = \frac{1}{T} \times \sum_{n=2}^{N} \mathbbm{1}(g_n==z_j)
			\times \mathbbm{1}(g_{n-1}\neq z_j)
		\end{aligned}
	\end{equation*}
	
	However, since gaze estimates are noisy, a majority rule over a buffered window
	is necessary to acknowledge transition into a new gaze zone. Algorithm
	\ref{alg:c03-GTF} details calculation of the glance frequency while accounting
	for noisy estimation. 

	\subsubsection{Glance Duration}
	Glance duration is a vector of size $L$, where each entry is a function of a unique gaze zone. Given a gaze zone, glance duration is the longest glance made towards the gaze zone of interest within time window $N$.
	Following the same process as in Algorithm \ref{alg:c03-GTF}, in addition to counting when
	transitions to new gaze zones occur, the start and end of each continuous glance can also be tracked. For gaze zone $z_j$, let $S_{z_j}$ be a $[N_j
	\times 2]$-matrix indicating the start and end index of $N_j=\textbf{C}_G(z_j)$
	continuous glance, where $\textbf{C}_G(z_j)$ is the number of continuous
	glances to $z_j$ as computed in Algorithm \ref{alg:c03-GTF}. Glance duration for
	each of the gaze zones, $z_j$, where $j \in \{1,2,...,L\}$ corresponds to the $j^{th}$ gaze zone in $Z$, is calculated as follows:
	\begin{equation*}
		\textit{Glance Duration } (z_j) =
		\begin{cases}
		\max\limits_{1\leq n\leq N_j} \left|\delta(S_{z_j}(n,:)\right| & \textit{if } N_j > 0 \\
		0 & \textit{if } N_j=0
		\end{cases} 
	\end{equation*}
	where $\delta(\bullet)$ is the difference operator.
	

	The final feature vector, $\vec{h}$, representing a scanpath then is made up of
	a combination of the above described descriptors. In particular, this paper will
	explore the benefits of representing a scanpath in three different ways: gaze
	accumulation alone, glance duration alone and glance duration plus glance frequency. 
	

	\subsection{Gaze Behavior Modeling}

	Consider a set of feature vectors $\vec{H} = \{\vec{h}_1, \vec{h}_2, \dots,
	\vec{h}_N\}$, and their corresponding class labels $Y = \{y_1, y_2, \dots,
	y_N\}$. In this paper, the class labels will be maneuvers: \textit{Left Lane Change},
	\textit{Right Lane Change}, \textit{Lane Keeping}. The gaze behaviors of respective
	events, tasks or maneuvers, are then modeled using an unnormalized multivariate normal
	distribution (MVN):
	\[
	M_{b}(\vec{h})=\exp \left(
	-\frac{1}{2}(\vec{h}-\vec{\mu_{b}})^T\Sigma_{b}^{-1}(\vec{h}-\vec{\mu_{b}})
	\right)
	\]
	
	where $b \in B = $ \{\textit{Left Lane Change}, \textit{Right Lane Change},
	\textit{Lane Keeping}\}, and $\mu_{b}$ and $\Sigma_{b}$ represent mean and
	covariance computed over the training feature vectors for the gaze behavior
	represented by $b$. One of the reasons for modeling gaze behavior in such a way is, given a
	new test scanpath descriptor, $\vec{h}_{test}$, we want to know how does it compare to the
	average scanpath computed for each gaze behavior in the training corpus. One
	possibility is to compute the euclidean distance between the average scanpath descriptor,
	$\mu_{b}$, and the test scanpath descriptor, $\vec{h}_{test}$, for all $b \in B$, and assign
	the label with the shortest distance. However, this assigns equal weight or
	penalty to every component in $\vec{h}$. The weights, however, should be a function of
	component as well as behavior under consideration. Therefore, we use the
	Mahalanobis distance, which assigns weights appropriately based on expected variance in the training data. Furthermore, by exponentiating the Mahalanobis distance to produce the unnormalized MVN, the range is mapped between $0$	and $1$. To a degree this can be used to asses the probability or confidence
	that a certain test scanpath represented by its descriptor, $\vec{h}_{test}$, belongs to a particular gaze
	behavior model. 
	
	
	\begin{algorithm}[!t]
		\SetKwInOut{Input}{input}\SetKwInOut{Output}{output}
		\Input{$G=[g_1,g_2,\dots,g_N]$ are noisy gaze estimates\newline
			$W$, a positive time window threshold for consistency check, $< N$
		}
		\Output{A vector, $\textit{\textbf{F}}_{G}$, of frequency of glances}
		\BlankLine
		\textit{LastGazeState}$=g_1$\\
		\For{$i \leftarrow W$ \KwTo $N$}{
			\If{$g_i \neq \textit{LastGazeState}$}{
				\If{\textit{Majority }($g_i== [g_{i-1} \cdots g_{i-W}])$}{
					$\textbf{C}_{G}(g_i)++$;\\
					$\textit{LastGazeState}=g_i;$
				}
			}
			i++;
		}
		$\textbf{F}_G = \frac{1}{T} \times \textbf{C}_G;$
		\caption{To compute a vector of glance frequencies given noisy estimates of
			gaze zones over time period $T$ with $N$ frames.}
		\label{alg:c03-GTF}
	\end{algorithm}
	
	\section{Experimental Design and Analysis}

	\subsection{Naturalistic Driving Dataset}
	A large corpus of naturalistic driving dataset is collected using an
	instrumented vehicle testbed. The vehicular tested is instrumented to
	synchronously capture data from camera sensors for looking-in and looking-out,
	radars, LIDARs, GPS and CAN bus. Of interest in this study are two camera
	sensors looking at the driver (i.e. one near the rearview mirror and another
	near the A-pillar) and one camera sensor looking-out in the driving direction. As the
	focus of this study is in driver gaze dynamics, the looking out view is only used to provide
	context for data mining. Using the same instrumented vehicle test, seven drivers
	of varying driving experience drove the car on different routes for an average
	of 40 minutes (see Table \ref{tab:c03-dataset-description}). Each drive
	consisted of some parts in urban settings, but mostly in freeway settings with
	multiple lanes. As the drivers are familiar with the area and were given the independence to design their own routes, the dataset contains natural glance behavior during driving maneuvers.

	\begin{table}[!t]
		\centering
		\caption{Description of analyzed on-road driving data.}
		{\footnotesize
			\begin{tabular}{|c|c|c|c|c|}
				\hline
				& Duration & \multicolumn{3}{c|}{No. of Events}\\
				\cline{2-5}
				Driver & Full drive & Left Lane & Right Lane & Lane\\
				ID & [min] & Change  & Change & Keeping\\
				\hline \hline
				1 & 52.10 & 9 & 5 & 20\\ 
				2 & 24.25 & 5 & 5 & 60\\ 
				3 & 28.13 & 5 & 4 & 50\\ 
				4 & 36.38 & 10 & 4 & 32\\ 
				5 & 39.20 & 10 & 4 & 45\\ 
				6 & 27.49 & 6 & 5 & 80\\ 
				7 & 37.50 & 5 & 5 & 46\\ 
				\hline
				All & 273.30 & 50 & 32 & 333\\
				\hline
			\end{tabular}
		}
		\label{tab:c03-dataset-description}
	\end{table}
		
	From the collected dataset of seven drivers, several types of annotations were
	done
	\begin{itemize}
		\item Gaze zone annotation of approximately equal number of samples with
		respect to the gaze-zones for all seven drivers. Each sample was annotated only
		when the annotator was highly confident that without ambiguity the sample falls
		into one of nine gaze-zone classes. Some annotated samples are from consecutive
		video frames while others are not. For a full description, the readers are
		referred to \cite{VoraIV2017}. Let's call this the Gaze-zone-dataset.
		\item Left and right lane change event annotations for all drivers. As a point
		of synchronization, for lane change events, when the vehicle tire is about to
		cross over into the other lane, it is marked at annotation and denoted as
		\textit{SyncF}. A 20-second window centered on \textit{SyncF}, makes up the
		event. At the time of training and testing, however, gaze dynamics is computed on a sliding 5-second window (see Section \ref{sec:c03-gaze-behavior-analysis}). Accumulated number of these events per driver and overall are given in
		Table \ref{tab:c03-dataset-description}. Let's call this the
		Lane-change-events-dataset.
		\item Lane keeping event annotations for all drivers. Lengthy stretches of lane keeping (as
		seen from looking-out camera) are broken into non-overlapping 5-second time window
		segments to create lane keeping events. Table \ref{tab:c03-dataset-description}
		contains the number of such events annotated and considered for the following analysis. Let's call this the
		Lane-keeping-events-dataset.
		\item Gaze zone annotation of every frame in the Lane-change-events-dataset. When annotating
		each of the 20-second continuous video segment, human annotators had to make a
		choice between one of the nine zones or \textit{unknown}. When ambiguous samples arose,
		the annotators used temporal information and outside context to make the call. \textit{Unknown}
		was highly discouraged to be used except for transitions between zones. Let's
		call this the Gaze-dynamics-dataset.
	\end{itemize}

	\begin{figure}[!t]
		\centering
		
		\includegraphics[width=0.95\linewidth]{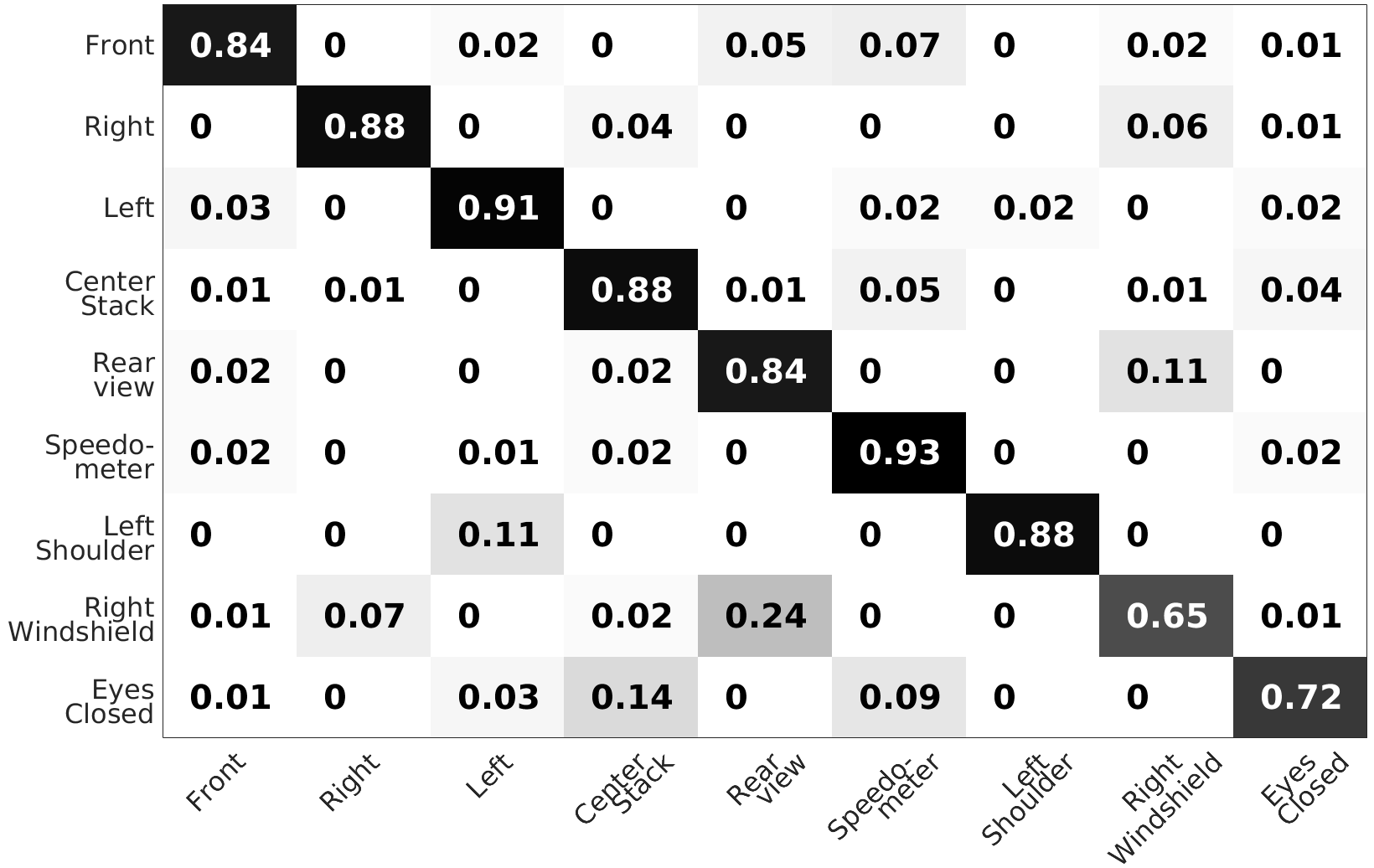}
		\caption{Evaluation of our gaze estimator on Gaze-zone-datase which is
			comprised of balanced samples with respect to gaze zones for each of the seven
			drivers. The confusion matrix is generated from a leave one driver out
			cross-validation, where the rows are true classes and columns are estimates. The
			rows as displayed may not sum to one because of \textit{Unknown}-class.}
		\label{fig:confumat_gaze-zone-dataset}
	\end{figure}
	\begin{figure*}[!t]
		\centering
		\begin{tabular}{cc}
			
			\includegraphics[width=0.45\linewidth]{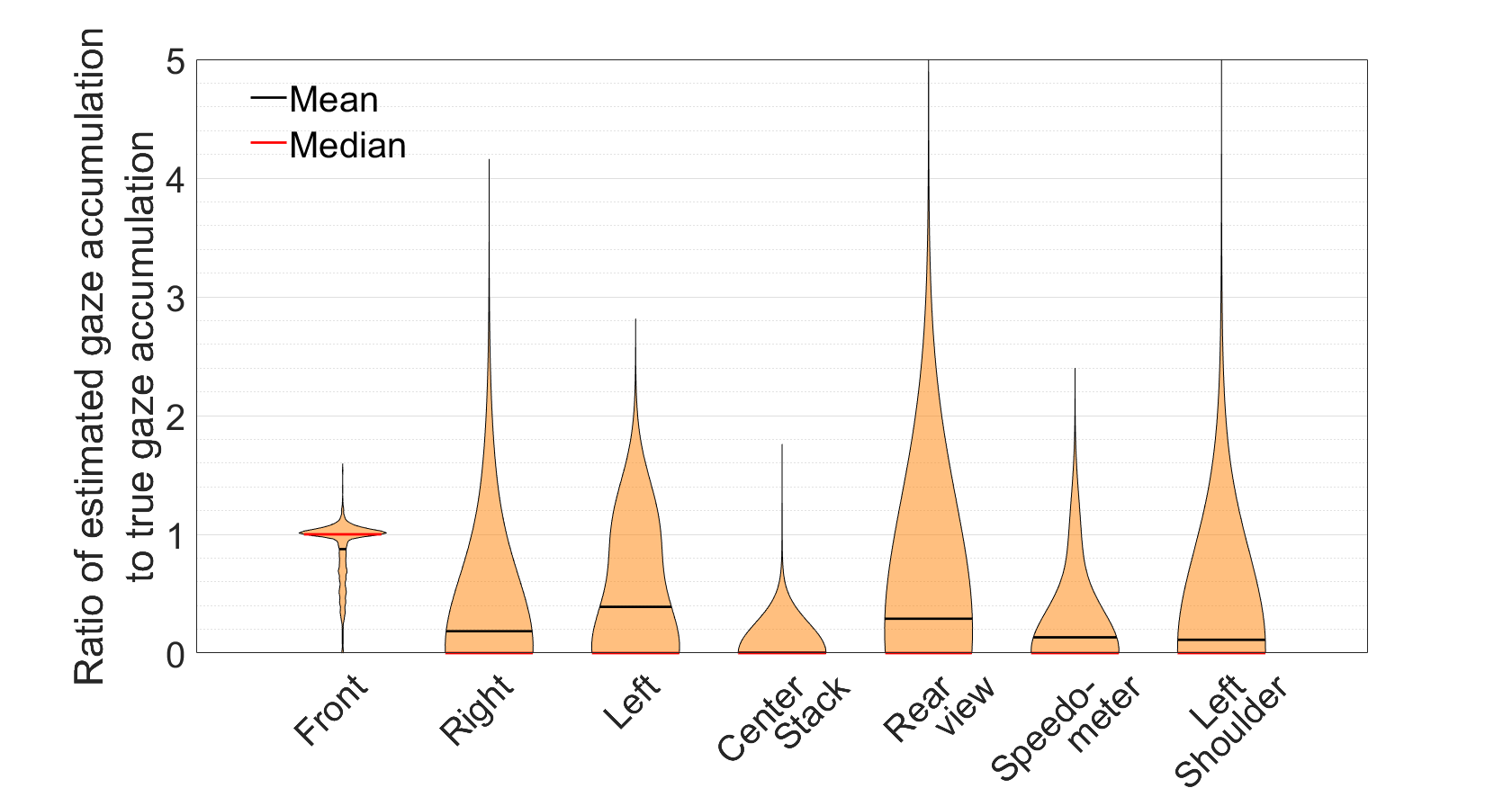}
			&
			
			\includegraphics[width=0.45\linewidth]{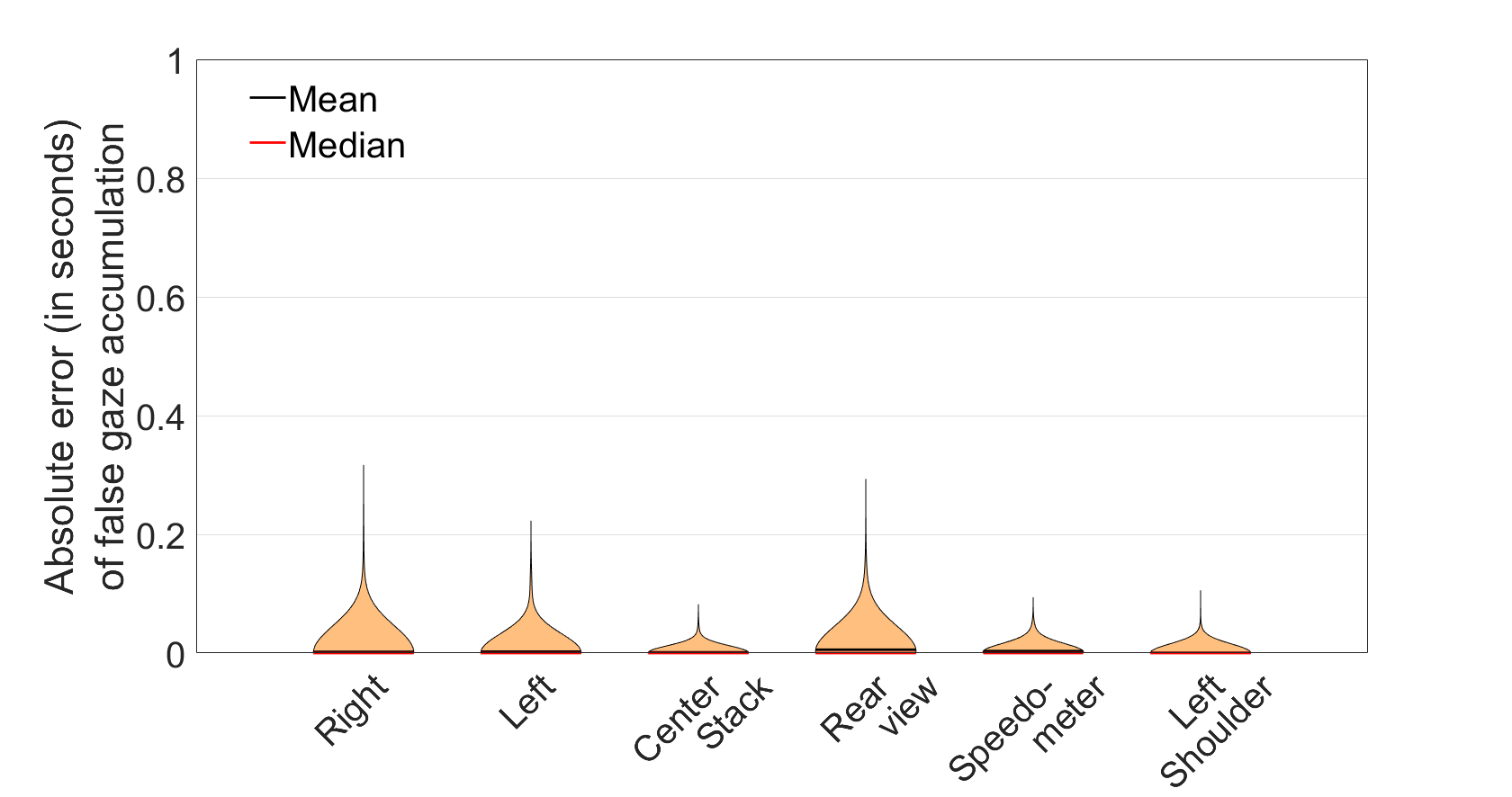}\\
			(a) Gaze accumulation ratio of true positives & (b) Gaze accumulation error of false positives
		\end{tabular}
		\caption{Performance evaluation of the gaze zone estimator is presented as a violin plot, which is a relative distribution of applying the following two metrics to all the samples in the Gaze-dynamics-dataset: (a) ratio of estimated to true gaze accumulation (Eq. \ref{eq:eval_metric_true_gaze_accumulation}) and (b) absolute error in estimated gaze accumulation due to false positives (Eq. \ref{eq:eval_metric_false_gaze_accumulation}). The width of the violin at respective values of the y-axis dictates the relative likelihood of the value for the gaze zone in the x-axis. Ideally, the width is largest for the ratio metric around one and for the absolute error metric around zero, meaning when true positives occur the durations of the estimated glances is close to durations of the true glances and when false positives occur the durations of those estimated glances are negligibly small.}
		\label{fig:durationErr}
	\end{figure*}
	\begin{figure*}[!t]
		\centering
		\begin{tabular}{ccc}
			\centering
			\includegraphics[width=0.25\linewidth]{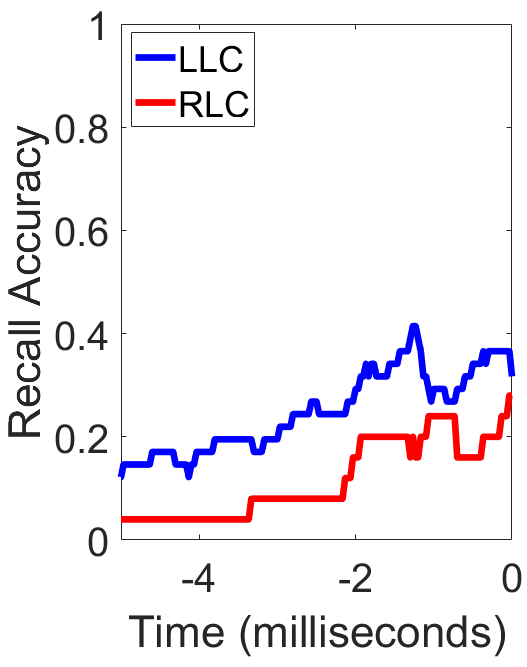} &
			\includegraphics[width=0.25\linewidth]{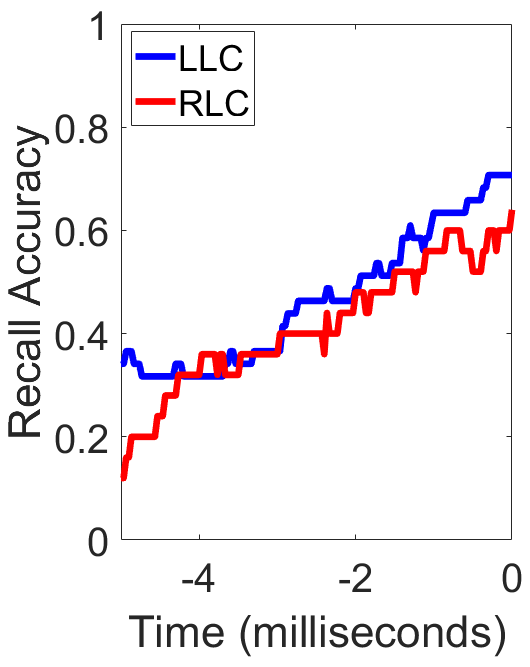} &
			\includegraphics[width=0.25\linewidth]{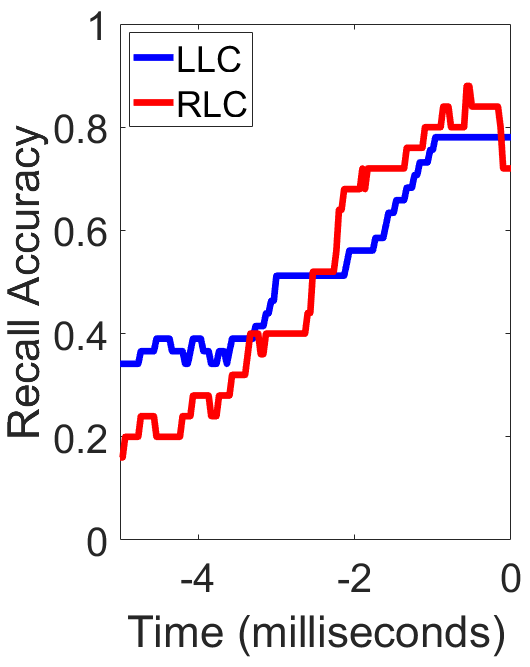} \\
			(a) Glance Duration + Frequency & (b) Glance Duration only & (c) Gaze
			accumulation\\
		\end{tabular}
		\caption{The recall accuracy of lane change prediction (averaged and
			cross-validated across all drivers in the naturalistic driving scenarios)
			continuously from -5 seconds to 0 seconds prior to lane change for three
			different combinations of spatio-temporal feature descriptors: (a) glance
			duration plus glance frequency, (b) glance duration only and (c) gaze
			accumulation only. LLC stands for left lane change and RLC stands for right lane change}
		\label{fig:recall_accuracy}
	\end{figure*}
	
	\subsection{Evaluation of Gaze Dynamics}
	Of the literary works listed in Table \ref{table:relatedWorks} which
	estimate gaze automatically, many of them output one of a number of gaze zones of
	interest. In those works, performance evaluation of their gaze estimator is presented in terms of a
	confusion matrix on what percent are correctly classified and what percent are
	misclassified with respect to the gaze zones. The advantage of such a
	presentation of evaluation is that when gaze zones are classified incorrectly, it
	shows what it is misclassified into and more often than not, the
	misclassification occurs in spatially neighboring zones (e.g. \textit{Front} and \textit{Speedometer}). 
	As for the dataset over which evaluation occurs, no guarantee is given that consecutive frames are annotated;
	in fact in \cite{TawariChenITSC2014}, annotations were done every 5 frames. As a point
	of comparison, Figure \ref{fig:confumat_gaze-zone-dataset} presents results of
	our gaze estimator (as described in Section \ref{sec:gaze_estimation}) on the
	Gaze-zone-dataset as a confusion matrix with a weighted accuracy of 83.5\% (i.e. accuracy is calculated per gaze zone and averaged  over all gaze zones) from leave one driver out cross-validation. 
	
	There are two important and often overlooked facts about this form of evaluation
	when considering the application of the gaze estimator on continuous video
	sequences and the interest is on semantics like gaze accumulation, glance
	durations and glance frequencies. First is the lack of evaluation on images or
	video frames where driver's gaze is in transition between two gaze zones. Since
	the gaze estimator is not explicitly trained to classify transition states, the
	gaze estimator is expected to ideally classify those transition instances into
	one of the two gaze zones that it is in transition between. 
	Second is the lack of 
	metrics to garner the effects of misclassification error on a continuous segment. For
	instance, consider the highest misclassification rate between front right
	windshield and rearview mirror seen in the confusion matrix in Figure
	\ref{fig:confumat_gaze-zone-dataset}. When does the misclassification occur? In
	the periphery of a continuous glance, in the middle of a glance or in transition
	between glances? Depending on the type of misclassification, it will affect
	glance duration and frequency calculation.
	
	The first limitation is addressed by creating the Gaze-dynamics-dataset. To address the latter limitation,
	this paper introduces two performance evaluation metrics for gaze dynamics with
	respect to gaze accumulation. One is the ratio of estimated gaze accumulation to
	true gaze accumulation per gaze zone:
	\begin{equation}
	\label{eq:eval_metric_true_gaze_accumulation}
	\textit{Relative ratio of $\hat{A}_G$ } (z_j) = 
	\begin{cases}
	\frac{\hat{A}_G(z_j)}{A_G (z_j)} & \textit{if } A_G(z_j) \neq 0\\
	0 & \textit{if }A_G(z_j) =0
	
	\end{cases}
	\end{equation}
	where $A_G(z_j)$ is the gaze accumulation calculated from ground truth
	annotation of gaze zones over a time period for gaze zone $z_j$ and
	$\hat{A}_G(z_j)$ is the gaze accumulation calculated from estimated gaze zones
	over a time period for gaze zone $z_j$. Note that in a given time window, only
	true positive gaze zone accumulation is considered with the first metric.
	Therefore, the second metric is designed to account for false gaze
	accumulations:
	
	\begin{equation}
	\label{eq:eval_metric_false_gaze_accumulation}
	\textit{Abs error of $\hat{A}_G$ } (z_j) = 
	\begin{cases}
	0 & \textit{if } A_G(z_j) \neq 0\\
	\hat{A}_G(z_j) & \textit{if }A_G(z_j) =0
	
	\end{cases}
	\end{equation}
	
	These new performance metrics are applied to the Gaze-dynamics-dataset, where
	each of the 20-second videos are broken into 5-second segments with up to
	4-second overlaps resulting in a total of 1312 samples. The performance is
	illustrated using a violin plot in Figure \ref{fig:durationErr}; a violin plot
	is a distribution of the output of the metrics over all the samples 
	in respective gaze zone classes. Ideally, the ratio
	metric (Eq. \ref{eq:eval_metric_true_gaze_accumulation}) is concentrated around 1, however, as
	seen in Figure \ref{fig:durationErr}, only the front gaze zone follows such a
	pattern. This result shows promise of accurately detecting attention versus
	inattention to the forward driving direction because the ratio of estimated to true gaze accumulation is highly concentrated around 1 for \textit{Front} gaze zone. Meanwhile, for other gaze zones, in majority of the samples, estimated gaze accumulation is less than the true gaze accumulation; this is mainly because glances towards these regions are significantly shorter in duration when compared to glance towards \textit{Front} and therefore more prone to noisy estimates.
	
	The second metric (Eq. \ref{eq:eval_metric_false_gaze_accumulation}) tries to answer the following questions: what happens when
	given a time period, ground truth annotations do not contain any annotations of a
	particular gaze zone but the gaze estimator produces false positives? Are the
	false positives sparse or significant in time? According to Figure
	\ref{fig:durationErr}b, the false gaze accumulations are small relative to the
	5-second window over which the gaze accumulation is calculated. Ideally, when calculating gaze accumulation over a time segment of estimated gaze zones, the ratio metric (Eq. \ref{eq:eval_metric_true_gaze_accumulation}) should be around one and the absolute error metric (Eq. \ref{eq:eval_metric_false_gaze_accumulation}) should be around zero, meaning when true positives occur the durations of the estimated glances is close to durations of the true glances and when false positives occur the durations of those falsely estimated glances are negligibly small.

	\subsection{Evaluation on Gaze Modeling}
	\label{sec:c03-gaze-behavior-analysis}
	All evaluations conducted in this study is done with a seven-fold cross
	validation; seven because there are seven different drivers as outlined in Table
	\ref{tab:c03-dataset-description}. With this setup of separating the training and
	testing samples, we explore the recall accuracy of the gaze behavior model in
	predicting lane changes as a function of time (Figure \ref{fig:recall_accuracy}).
	
	Training occurs on the 5-second time window before \textit{SyncF} as represented
	by the events in Table \ref{tab:c03-dataset-description}. Note that, manually
	annotated gaze zones are used to compute the spatio-temporal features used to
	train the lane change models whereas estimated gaze zones are used to train the
	lane keeping model. In testing, however, only estimated gaze-zones are used to
	compute spatio-temporal features.
	
	\begin{figure*}[!t]
		\centering
		\begin{tabular}{cc}
			\centering
			%
			%
			%
			\includegraphics[width=0.42\linewidth]{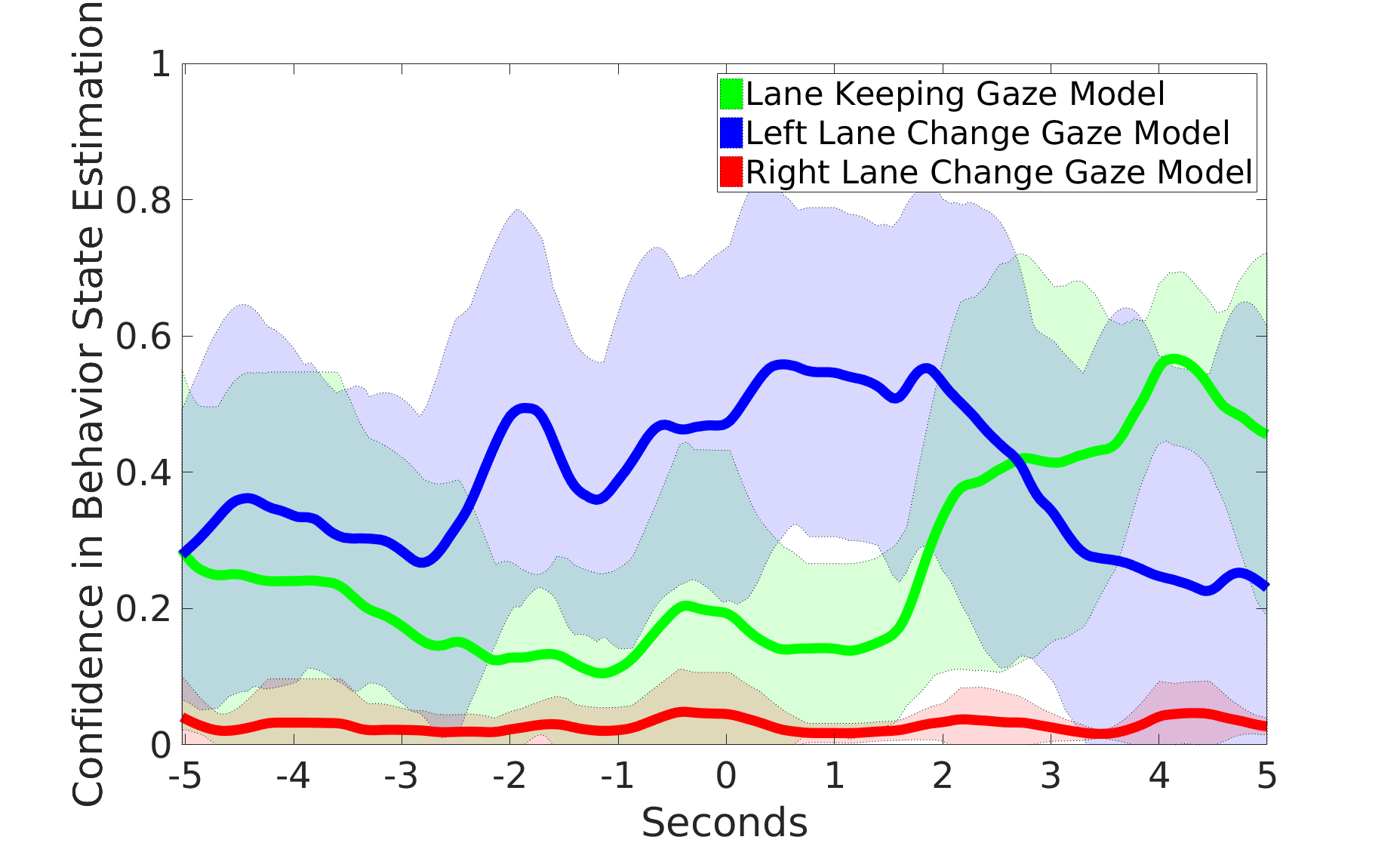} & 
			\includegraphics[width=0.42\linewidth]{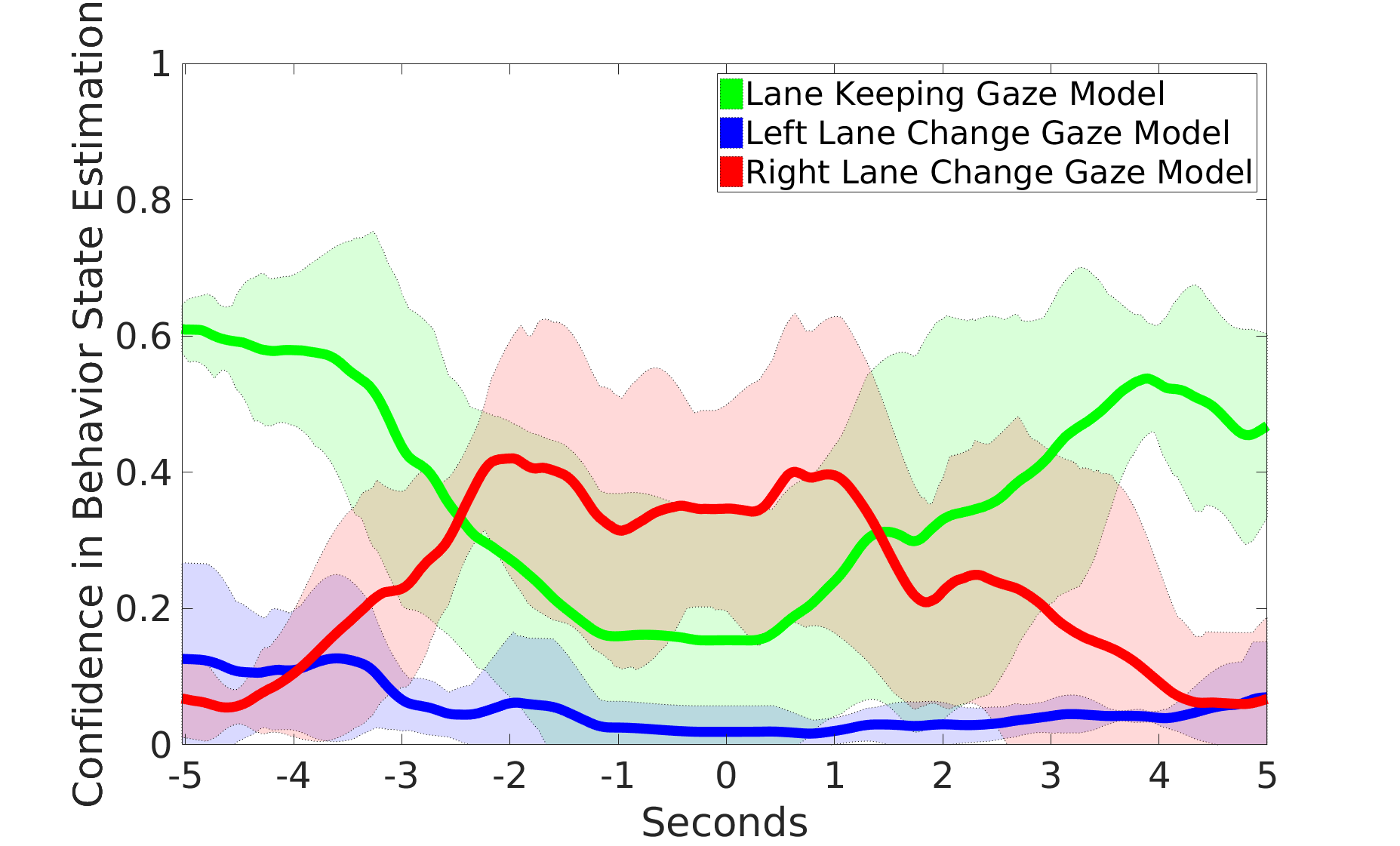}\\
			(a) Left Lane Change Events (DriverID 2) & (b) Right Lane Change events
			(DriverID 2)\\
			\includegraphics[width=0.42\linewidth]{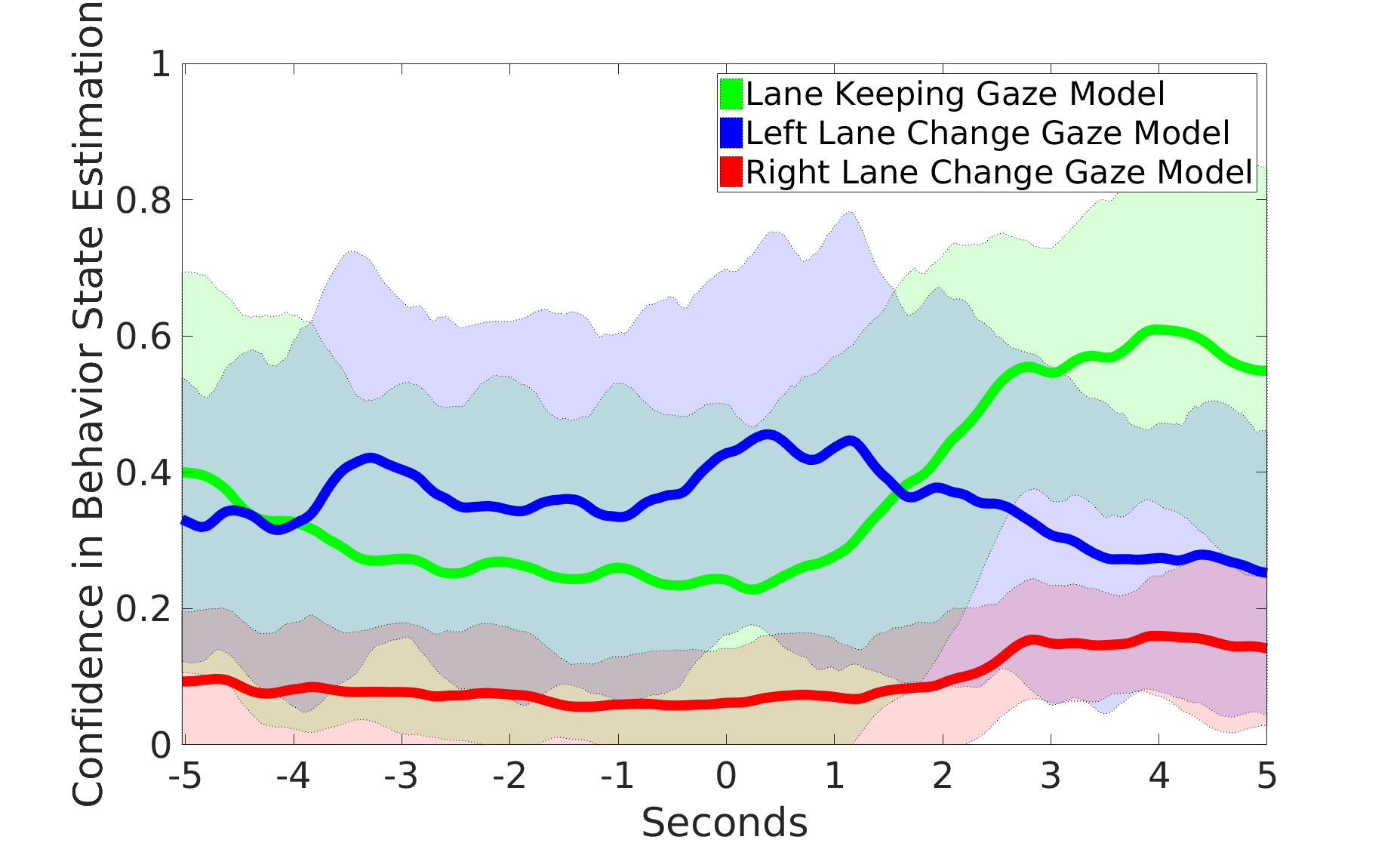} & 
			\includegraphics[width=0.42\linewidth]{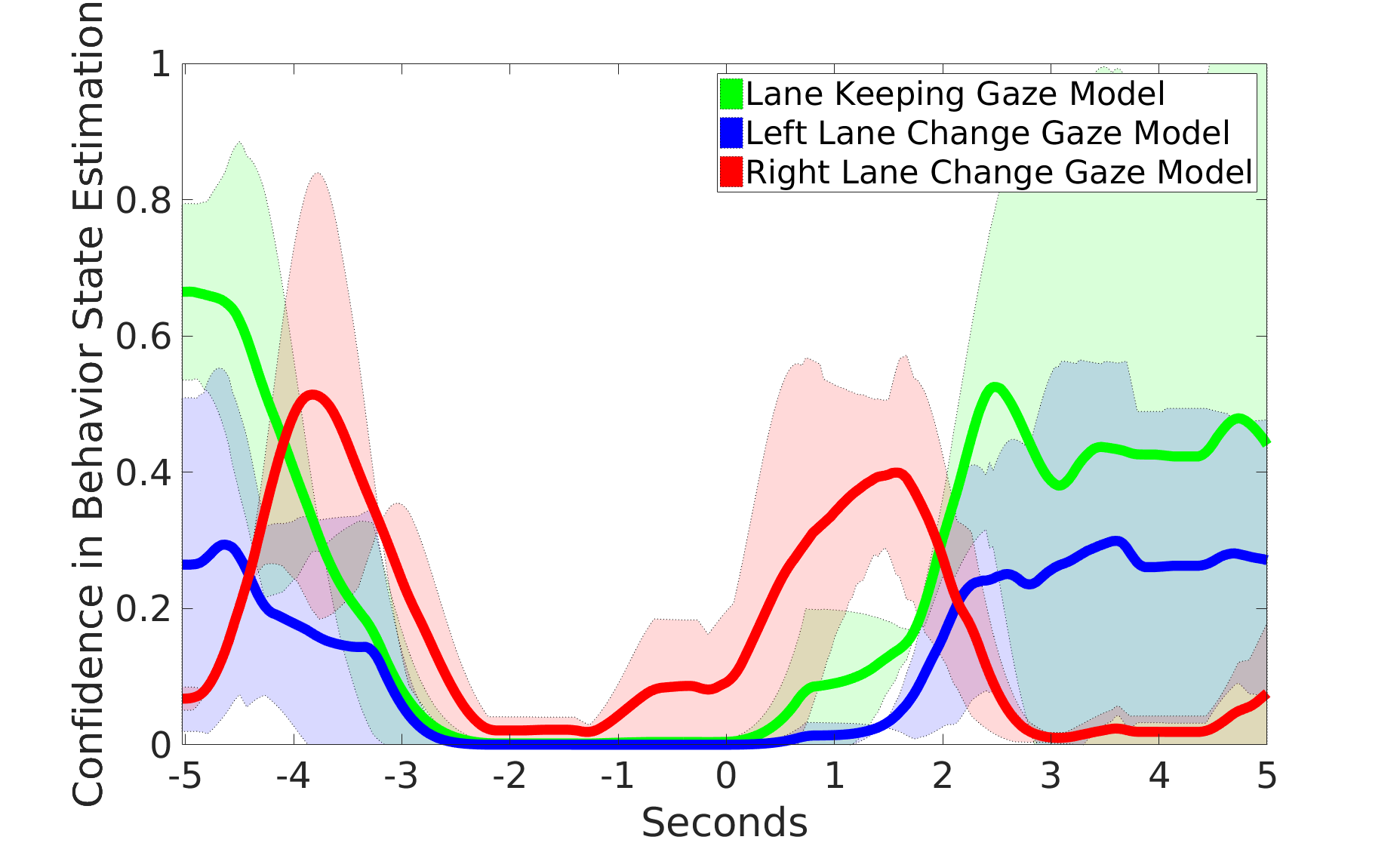}\\
			(c) Left Lane Change Events (Driver ID 5)& (d) Right Lane Change events
			(DriverID 5)\\
		\end{tabular}
		\caption{Illustrates variations in fitness of the three models (i.e.
			\textit{Left lane change}, \textit{Right lane}) during left and right lane
			change maneuvers for two different drivers, where mean and standard deviation are depicted with solid line and semitransparent shades, respectively.}
		\label{fig:c03-model-confidence}
		\vspace{-5mm}
	\end{figure*}
	
	At testing time, we want to test how early the gaze behavior models are able to
	predict lane change. Therefore, starting from 5-seconds before $\textit{SyncF}$
	sequential samples with $\frac{1}{30}$ of a second overlap are extracted up to
	5-seconds after \textit{SyncF}; note that the time window at 5-seconds before
	the \textit{SyncF} encompasses data from 10 seconds before the \textit{SyncF} up
	to 5-seconds before the \textit{SyncF}. Each of the samples are tested for
	fitness across the three gaze behavior models, namely models for \textit{left
		lane change}, \textit{right lane change} and \textit{lane keeping}. The sample is
	assigned the label based on the model which procures the highest fitness score
	and if the label matches the true label the sample is considered a true
	positive. Note that each test sample is associated with a time index of where it
	is sampled from with respect to \textit{SyncF}. By gathering samples at the same
	time index with respect to \textit{SyncF}, recall value at a given time index is calculated by dividing the number of true positives by the total number of positive samples. 
	
	When calculating recall values, true labels of samples were remapped from three
	classes to two classes; for instance, when computing recall values for left lane
	change prediction, all right lane change events and lane keeping events were
	considered negatives samples and only the left lane change events are considered
	positive samples. Similar procedure is observed for computing recall values for
	right lane change prediction. Figure \ref{fig:recall_accuracy} shows the
	development of the recall values for both left and right lane change prediction
	continuously from -5 seconds prior to \textit{SyncF} up to 0 milliseconds prior
	to \textit{SyncF} for three different combinations of features (i.e. glance
	duration plus frequency, glance duration only and gaze accumulation only) and
	for two events (i.e. left lane change and right lane change). As expected the
	recall curves rise in accuracy the closer in time to the lane change event. Also
	expected is the performance difference with respect to the spatio-temporal
	features; whereas modeling with gaze accumulation alone achieves above 75\% accuracy at 1000 milisecond prior to lane change, modeling with glance duration alone and glance duration plus frequency achieves about 60\% and 40\% accuracy, respectively. One possible reason for the stark difference in performance when using
	gaze accumulation versus glance duration and frequency is the latter may vary
	across drivers more than the former. For example, one driver may exhibit short
	glances with high frequency whereas another driver may make long glances with
	low frequency. However, gaze accumulation neatly maps the differences in gaze
	behavior to one ``attention'' allocation domain and therefore gives the best
	performance under given modeling methods and dataset.

	Lastly, in Figure \ref{fig:c03-model-confidence}, we
	illustrate the fitness or confidence of the learned models around left and
	right lane change maneuvers for two different drivers. The figure shows mean
	(solid line) and standard deviation (semitransparent shades) of three models
	(i.e. \textit{left lane change}, \textit{right lane change}, \textit{lane keeping})
	using the events from naturalistic driving dataset described in Table
	\ref{tab:c03-dataset-description}. The model confidence statistics are plotted 5
	seconds before and after the lane change maneuver, where time of 0 seconds
	represents when the vehicle is about to change lanes. Interestingly, even though
	early versus late peaks of the appropriate model can be uniquely different across drivers and maneuvers, 
	the satisfactory separation of the lane change models to lane keeping model and the spread in dominance
	of the correct model shows promise in modeling driver behavior using gaze dynamics to anticipate activities and
	maneuvers.
	
	
	\section{Concluding Remarks}
	In this study, we explored modeling driver's gaze behavior in order to predict
	maneuvers performed by drivers, namely left lane change, right lane change and
	lane keeping. The particular model developed in this study features three major
	aspects: one is the spatio-temporal features to represent the gaze dynamics,
	second is in defining the model as the average of the observed instances, third
	is in the design of the metric for estimating fitness of model. Applying this
	framework in a sequential series of time windows around lane change maneuvers,
	the gaze models were able to predict left and right lane change maneuver with an
	accuracy above 75\% around 1000 milliseconds before the maneuver.
	
	The overall framework, however, is designed to model driver's gaze behavior for
	any tasks or maneuvers performed by driver. In particular, the spatio-temporal
	feature descriptor composed of gaze accumulation, glance duration and glance
	frequency are powerful tools to capture the essence of recurring driver gaze
	dynamics. To this end, there are multiple future directions in site. One is to
	quantitatively define the relationship between the time window from which to
	extract those meaningful spatio-temporal features and the task or maneuvers
	performed by driver. Other future directions are in exploring and comparing
	different temporal modeling approaches and generative versus discriminative models.
	

	\section*{Acknowledgment}
	The authors would like to thank the reviewers and the 
	editors for their constructive and encouraging feedback, 
	and their colleagues at the Laboratory of Intelligent and 
	Safe Automoilbles (LISA). The authors gratefully
	acknowledge the support of UC Discovery Program and 
	industry partners, especially Fujitsu Ten and 
	Fujitsu Laboratories of America.
	
	
	\bibliographystyle{IEEEtran}
	\bibliography{FINAL_VERSION}
	\vspace{-20mm}
	
	\begin{IEEEbiography}[{\includegraphics[width=1in,height=1.25in,clip,keepaspectratio]{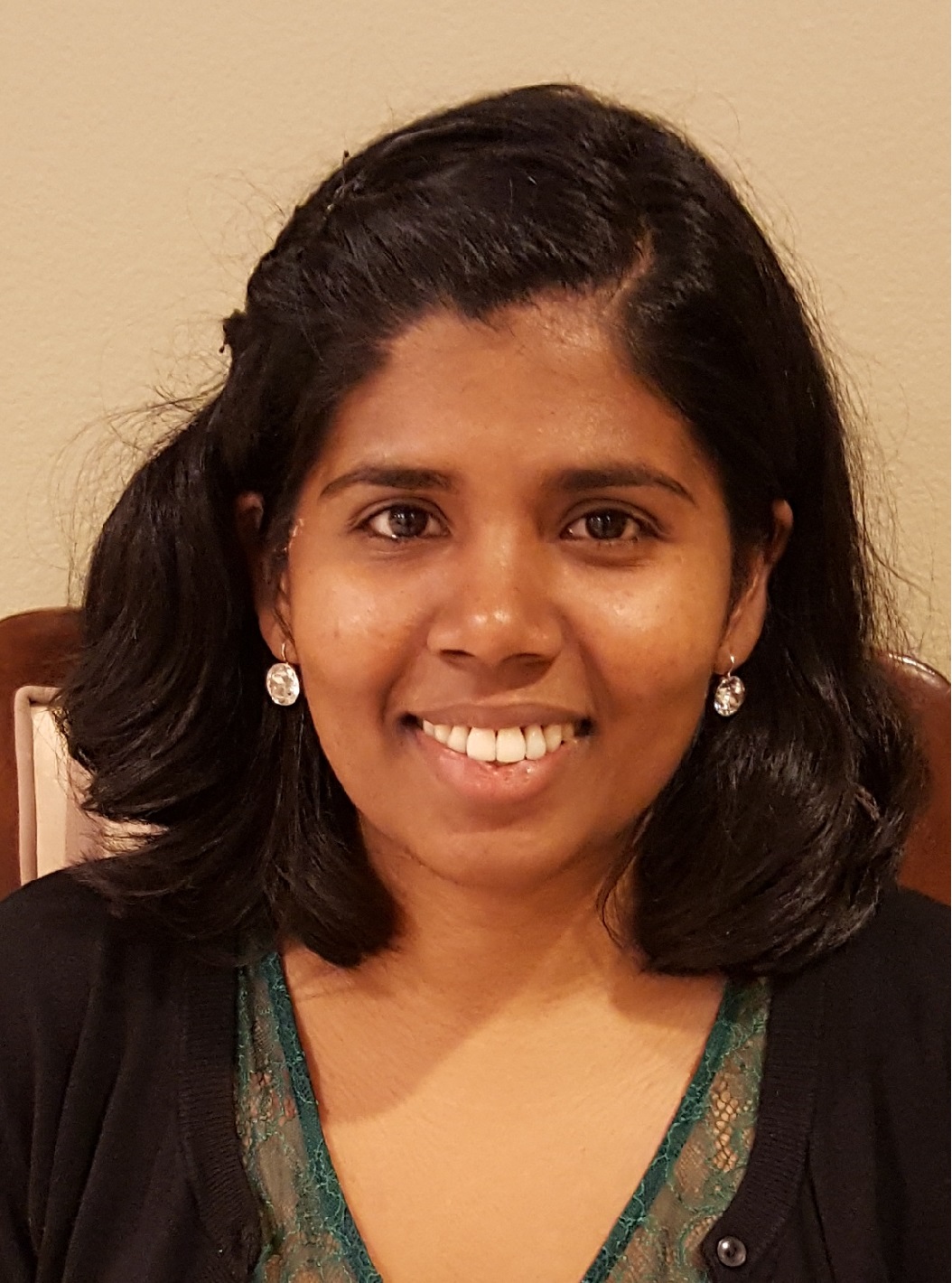}}]{Sujitha Martin}
		received the BS degree in Electrical Engineering from the California Institute
		of Technology in 2010 and the MS and Ph.D degree in electrical engineering from
		the University of California, San Diego (UCSD) in 2012 and 2016, respectively.
		She is currently a research scientist at Honda Research Institute USA. Her research interests
		are in machine vision and learning, with a focus on human-centered,
		collaborative, intelligent systems and environments. She helped organize two
		workshops on analyzing faces at the IEEE Intelligent Vehicles Symposium (IVS
		2015, 2016) and the first Women in Intelligent Transportation Systems (WiTS)
		meet and greet networking event at the IEEE IVS (2017). She is recognized as one of top
		female graduates in the fields of electrical and computer engineering and
		computer science at Rising Stars 2016 hosted by Carnegie Mellon University.
		\vspace{-20mm}
	\end{IEEEbiography}
	
	
	\begin{IEEEbiography}[{\includegraphics[width=1in,height=1.25in,clip,keepaspectratio]{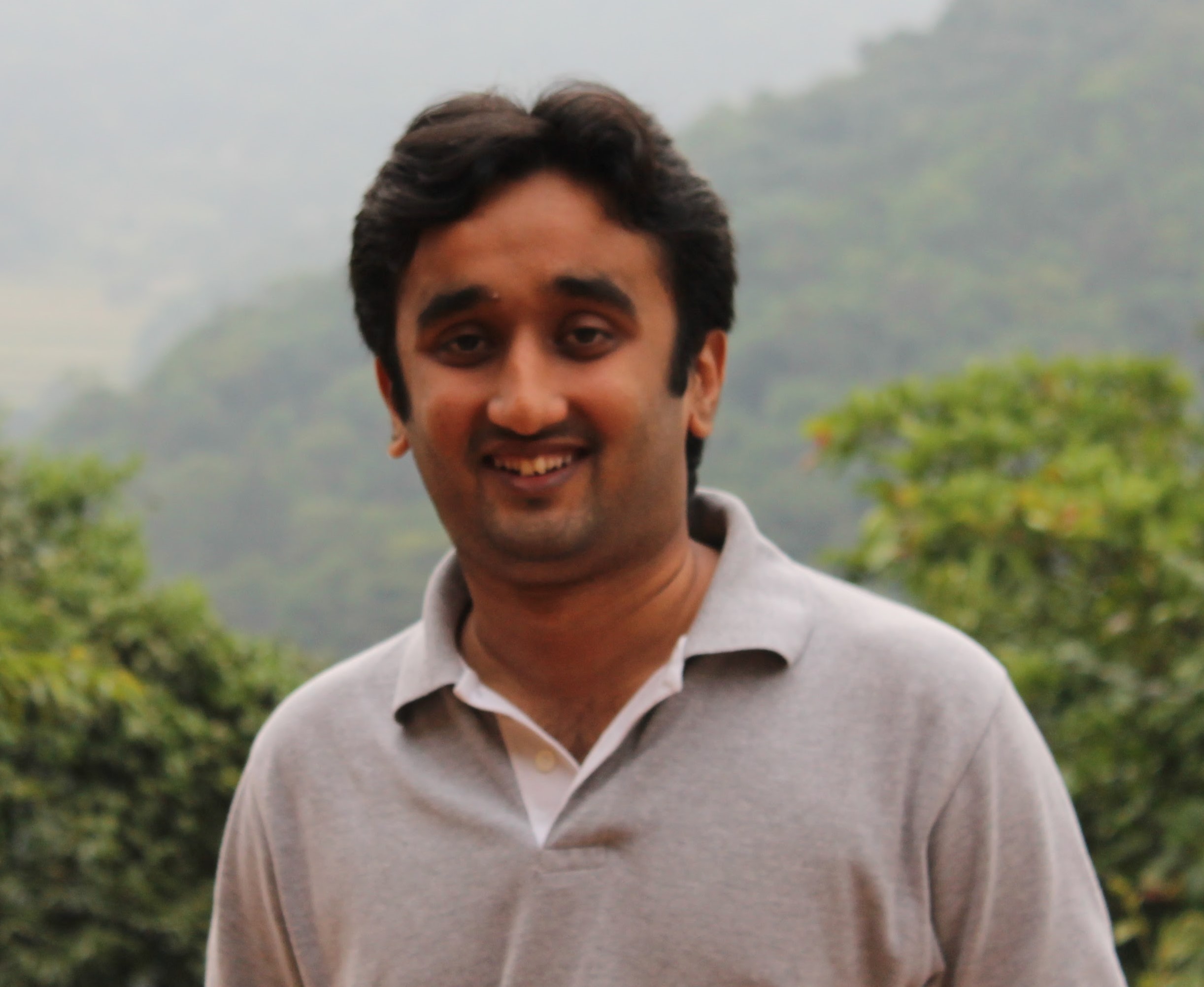}}]{Sourabh
			Vora} received his BS degree in Electronics and Communications Engineering (ECE) from Birla Institute of Technology and Science (BITS) Pilani - Hyderabad Campus. He received his MS degree in Electrical and Computer Engineering (ECE) from University of California, San Diego (UCSD) where he was associated with the Computer Vision and Robotics Research (CVRR) Lab. His research interests lie in the field of Computer Vision and Machine Learning. He is currently working as a Computer Vision Engineer at nuTonomy, Santa Monica.
		\vspace{-20mm}
	\end{IEEEbiography}
	
	\begin{IEEEbiography}[{\includegraphics[width=1in,height=1.25in,clip,keepaspectratio]{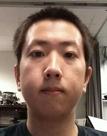}}]{Kevan
			Yuen}
		Kevan Yuen received the B.S. and M.S. degrees in electrical and computer
		engineering from the University of California, San Diego, La Jolla. During his
		graduate studies, he was with the Computer Vision and Robotics Research
		Laboratory, University of California, San Diego. He is currently pursuing a PhD
		in the field of advanced driver assistance systems with deep learning, in the
		Laboratory of Intelligent and Safe Automobiles at UCSD.
		\vspace{-20mm}
	\end{IEEEbiography}
	
	\begin{IEEEbiography}[{\includegraphics[width=1in,height=1.25in,clip,keepaspectratio]{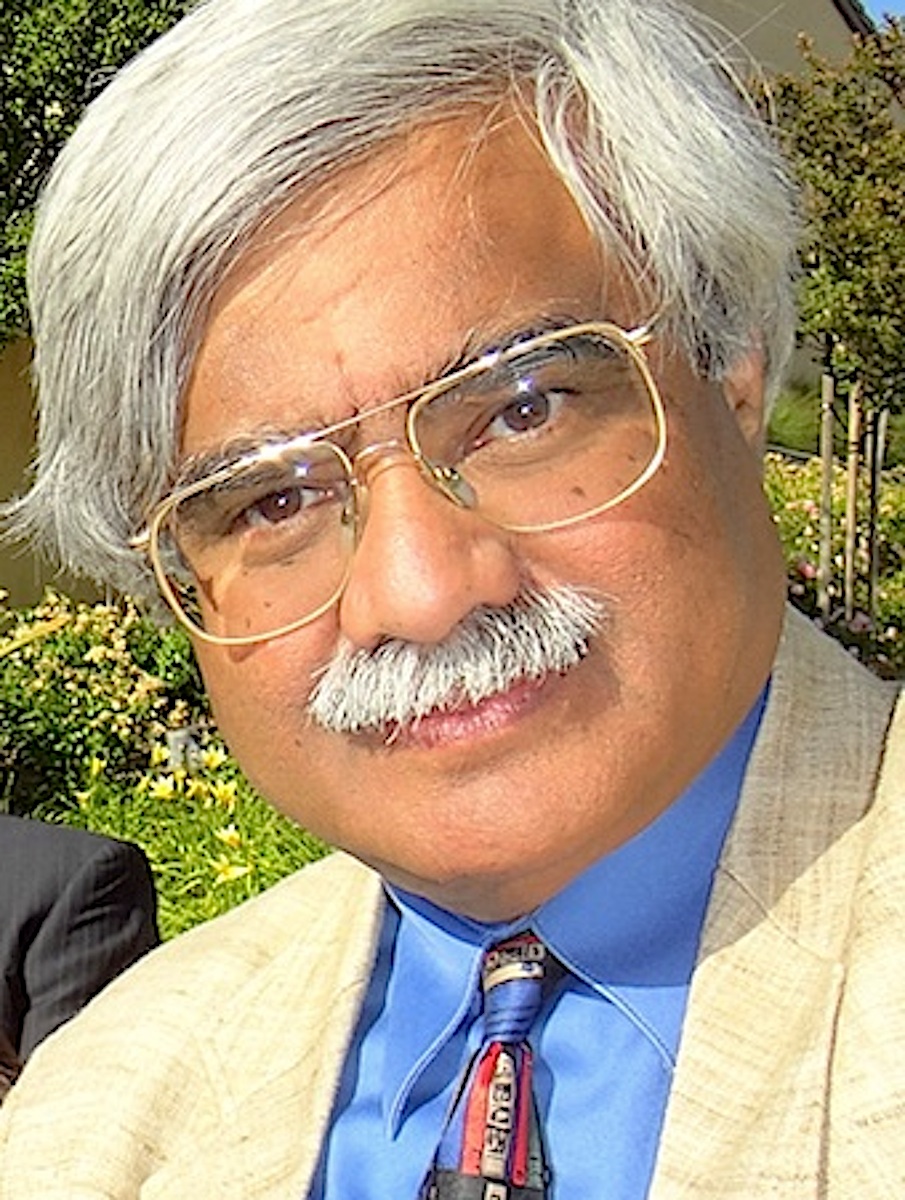}}]{Mohan
			Manubhai Trivedi} is a Distinguished Professor of and the founding director of
		the UCSD LISA: Laboratory for Intelligent and Safe Automobiles, winner of the
		IEEE ITSS Lead Institution Award (2015). Currently, Trivedi and his team are
		pursuing research in distributed video arrays, human-centered self-driving vehicles, human-robot interactivity, machine vision, sensor fusion, and active learning. Trivedi's team has played key roles in several major research initiatives. Some
		of the professional awards received by him include the IEEE ITS Society's
		highest honor ``Outstanding Research Award'' in 2013, Pioneer Award (Technical
		Activities) and Meritorious Service Award by the IEEE Computer Society, and
		Distinguished Alumni Award by the Utah State University and BITS, Pilani. Three of his students
		were awarded ``Best Dissertation Awards'' by professional societies and 20+ ``Best'' or ``Honorable Mention'' awards at international conferences. Trivedi is a Fellow of the IEEE, IAPR and SPIE. Trivedi regularly serves as a consultant to industry and government agencies in the U.S., Europe,
		and Asia.
	\end{IEEEbiography}
	
\end{document}